\def\year{2021}\relax
\documentclass[letterpaper]{article} %
\usepackage{aaai21}  %
\usepackage{times}  %
\usepackage{helvet} %
\usepackage{courier}  %
\usepackage[hyphens]{url}  %
\usepackage{graphicx} %
\def\UrlFont{\rm}  %
\usepackage{natbib}  %
\usepackage{caption} %
\usepackage[switch]{lineno}

\usepackage{xspace}

\def\OurExp{{Ours}\xspace}
\newcommand{\myparagraph}[1]{{\vspace{0.5em} \noindent \bf #1}}

\def\eg{\textit{e.g.}}
\def\ie{\textit{i.e.}}
\def\wrt{\textit{w.r.t.\ }}
\def\vs{\textit{vs.\ }}

\usepackage{graphicx}  %
\usepackage{multirow}
\usepackage{adjustbox}
\usepackage{tabu}
\usepackage{amssymb}
\usepackage{booktabs}
\usepackage{xcolor}
\usepackage{amsmath}
\usepackage{footnote, eucal}
\makesavenoteenv{table}
\def\xinyu{\textcolor{black}}

\title{Diverse Knowledge Distillation for End-to-End Person Search}
\author{
Xinyu Zhang$ ^{1,2}$, 
~~
Xinlong Wang$ ^{2}$,
~~
Jia-Wang Bian$ ^{2}$,
~~
Chunhua Shen$ ^{2,3}$
~~
Mingyu You$ ^{1}$
\\[0.25cm]
\rm 
$ ^1$ Tongji University, China
~~
$ ^2$ The University of Adelaide, Australia
~~
$ ^3$ Monash University, Australia
}

\begin{document}

\maketitle

\begin{abstract}
Person search aims to localize and identify a specific person from a gallery of images.
Recent methods can be categorized into two groups, \ie, two-step and end-to-end approaches.
The former views person search as two independent tasks and achieves dominant results using separately trained person detection and re-identification (Re-ID) models.
The latter performs person search in an end-to-end fashion.
Although the end-to-end approaches yield higher inference efficiency, they largely lag behind those two-step counterparts in terms of accuracy.
In this paper, we argue that the gap between the two kinds of methods is mainly caused by the Re-ID sub-networks of end-to-end methods.
To this end, we propose a simple yet strong end-to-end network with diverse knowledge distillation to break the bottleneck.
We also design a spatial-invariant augmentation to assist model to be invariant to inaccurate detection results.
Experimental results on the CUHK-SYSU and PRW datasets demonstrate the superiority of our method against existing approaches---it achieves on par accuracy with state-of-the-art two-step methods while maintaining high efficiency due to the
single joint model.
{   
    \def\UrlFont{\sf}
    \def\UrlFont{\rm\small\ttfamily}
Code is available at: \url{https://git.io/DKD-PersonSearch} 
}

\end{abstract}

\section{Introduction}
Person search~\cite{xu2014person} is a recent hot topic in computer vision.
The goal is to find a target person in a gallery of images,
and hence involves both person detection and re-identification (Re-ID).
Compared with solo Re-ID problem, person search 
faces 
greater
challenges posed by detection, \eg, inaccurate and misaligned bounding boxes may lead to poor identification.

\begin{figure}[t!]
\centering
\includegraphics[trim =0mm 0mm 0mm 0mm, clip, width=1\linewidth]{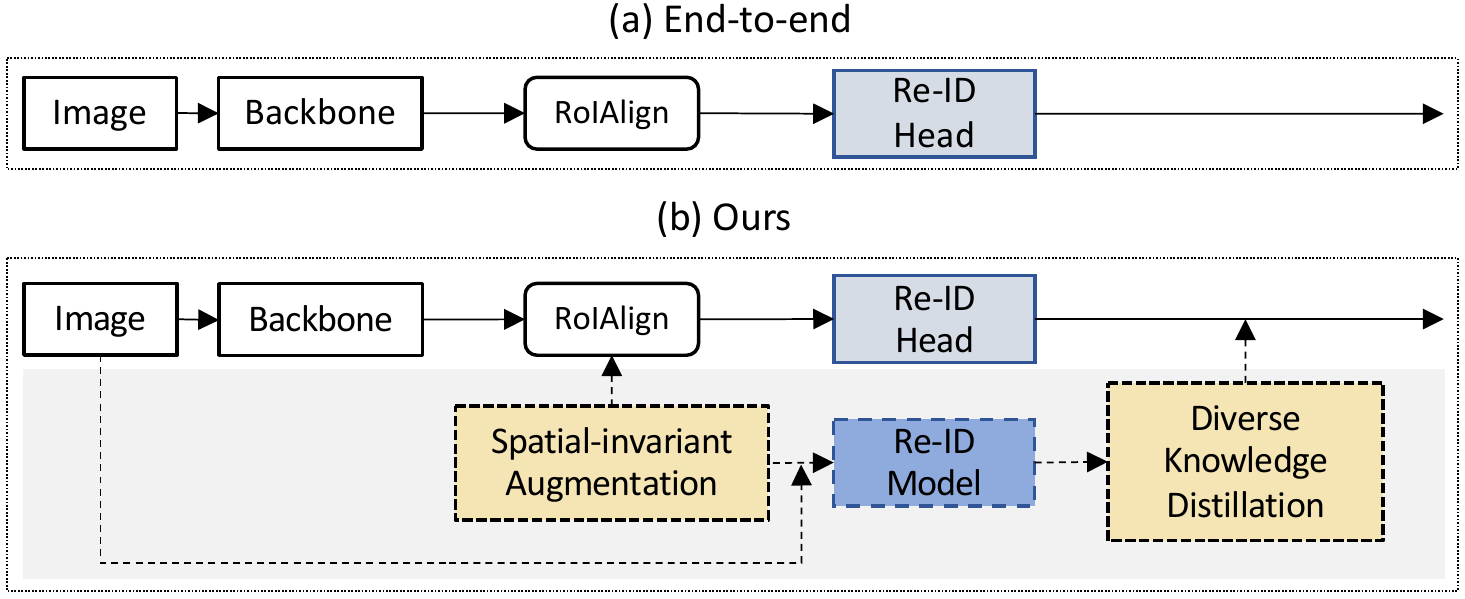}
\caption{Structure comparison between (a) the plain end-to-end model and (b) our method for person search.
Dashed lines are only involved during training.
Our method maximizes the usage of guidance from the external Re-ID model via the proposed terms---see dashed boxes.
}
\label{fig:figure1}
\end{figure}

Existing approaches for person search can be classified into two categories, \ie, two-step strategy~\cite{han2019re,lan2018person,han2019re,wang2020tcts,dong2020instance} and end-to-end strategy~\cite{xiao2017joint,xiao2019ian,liu2017neural,chang2018rcaa,he2018end,yan2019learning,munjal2019query,munjal2019knowledge,zhong2020robust,dong2020bi,chen2020norm}.
Two-step approaches tackle the detection and Re-ID problem independently, in which the detection model crops the candidate persons from raw images, and the Re-ID model 
extracts features for matching.
These methods have shown high matching accuracy,
but lag behind those end-to-end methods in efficiency
due to the cascaded processing.

Thanks to the shared backbone for detection and Re-ID, the end-to-end methods boost the speed significantly.
However, the performance 
is limited due to
the conflict objectives of these two components in the joint learning framework~\cite{chen2020norm}.
Specifically, the detection head tends to find a common embedding space for all pedestrians,
while the Re-ID head distinguishes different pedestrians with unique representations.
Besides, feature maps of the proposals involve redundant context information~\cite{zhu2019deformable,dong2020bi}, 
as they are pooled from the middle layer in the deep network, in which one pixel 
perceives a large area in
the raw image.
These issues consequently result in the inferior Re-ID accuracy.

\begin{figure*}[t!]
\centering
\includegraphics[trim =0mm 0mm 0mm 0mm, clip, width=0.9\linewidth]{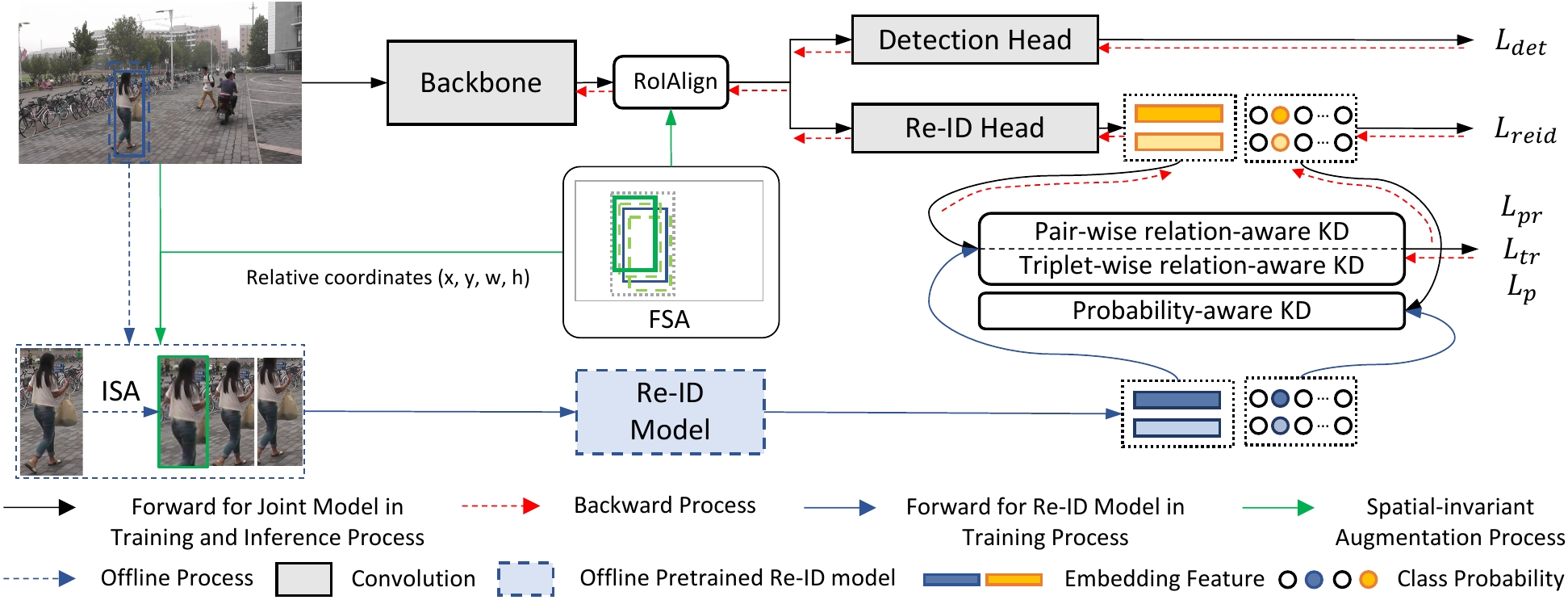}
\caption{Overview of our person search framework. We first pre-train the external Re-ID model using our proposed image-level spatial-invariant augmentation (ISA).
Then given the ground truth proposals, we use our proposed feature-level spatial-invariant augmentation (FSA) to obtain shifted bounding boxes in both images and feature maps.
The internal Re-ID head and external Re-ID model take the shifted boxes in feature maps and images as input, respectively.
Finally, we use three kinds of knowledge distillation (KD) to let the internal Re-ID head mimic the capability of the external Re-ID model,
including the probability-aware KD and the proposed pair-wise as well as triplet-wise relation-aware KD.
}
\label{fig:figure2}
\end{figure*}

In this paper, we boost the performance of end-to-end models towards efficient yet accurate person search.
We first analyze the gap of both detection and Re-ID accuracy separately
between two-step and end-to-end solutions.
As shown in Table~\ref{tab:Re-ID_importance}, the detection results are nearly the same in both settings,
while 
there exists a large gap between final results (\ie, Re-ID accuracy).
Therefore, we argue that the Re-ID head is the main bottleneck of end-to-end models.
We 
conjecture the reasons are twofold:
(1) Two subtasks have conflicts when learning the shared features, which causes the learned features less discriminative for distinguishing fine-grained persons;
(2) The Re-ID head highly overfits the bounding box predictions, and 
inaccurate
detection bounding box makes the matching fail, even it is a correct one (\ie, IoU with ground truth is above 0.5).

To capitalize on our findings,
we propose a simple yet effective person search framework
for boosting the performance of the Re-ID head within the joint model\footnote{Joint model is the alias of the end-to-end model in our paper.}.
The improvements come from two aspects.
The first one is the enhancement on the output side with diverse knowledge distillation scheme.
Since the Re-ID model in two-step approaches is trained on the person patches cropped
from the raw images with
ground truth bounding boxes, it can capture the discriminate features unaffected by the detection task. 
Inspired by knowledge distillation (KD)~\cite{hinton2015distilling}, we make use of a separately trained Re-ID model to provide the superior prior knowledge to the Re-ID head.
Apart from the traditional KD on the probability distribution, we design two relation-aware mimicking loss terms to make the Re-ID head fully emulate the sample relationships revealed by the well trained Re-ID model. 
The second one is a spatial-invariant augmentation mechanism on the input of the Re-ID head and the guiding Re-ID model. 
Instead of directly feeding ground truth person patches to the Re-ID head/model, we manually shift the cropping area of the proposals in the RoIAlign layer
to mimic inaccurate detection results. 
To make the Re-ID head insensitive to this spatial disturbance, 
we present another KD guidance on the shifted proposals to minimize the difference between the probabilities produced by the Re-ID head and the Re-ID model. 
In this way, the Re-ID head is encouraged to improve the robustness on the detection variances.

Although external guidance is introduced, 
our method remains a lightweight framework since 
only the end-to-end model is reserved in the inference stage.
Our contributions can be summarized as:
\begin{itemize}
\itemsep -.051cm
\item We experimentally find that the main bottleneck of the end-to-end person search is the inferior Re-ID head. 
\item We propose a novel person search framework
to enhance the Re-ID head via the diverse knowledge distillation and the spatial-invariant augmentation.
\item We demonstrate the efficiency and efficacy of our our method with comprehensive experiments,
and show state-of-the-art 
results on the CUHK-SYSU and PRW datasets.
\end{itemize}

\subsection{Related Work}
\myparagraph{Person search.}
Existing methods in person search can be divided into two types, \ie, two-step and end-to-end methods, according to performing the detection and Re-ID separately or jointly.
Two-step methods train detection and Re-ID models separately and then combine them together in sequence~\cite{zheng2017person,lan2018person,chen2018person,han2019re,chen2018person}.
\cite{zheng2017person} evaluate various combinations of detectors and identifiers and propose a confidence weighted similarity method for person matching.
\cite{lan2018person} study the multi-scale matching problem and propose a cross-level semantic alignment to address the problem.
\cite{chen2018person} enrich the feature representations by a mask-guided two-stream model.
\cite{han2019re} point out that the detected bounding boxes are sub-optimal for the Re-ID task and develop an ROI transform layer to refine the boxes, in which the gradient flow is from the identifier.
Recent work~\cite{wang2020tcts} consider the consistency requirements between detection and Re-ID,
and propose an identity-guided query detector and a detection results adapted re-ID model.  
These methods show that the two-step strategy can effectively alleviate the 
dilemma 
of detection and Re-ID and achieve high performance.
Because of the separate processing of the two subtasks, these methods can be slow.

End-to-end methods are proposed to improve the efficiency by training a joint model.
Among these methods, \cite{chang2018rcaa,liu2017neural} directly search the person from the uncropped images by decision making or reinforcement learning.
~\cite{zhong2020robust} address the misalignment issue and conduct robust partial matching.
\cite{chang2018rcaa,yan2019learning,dong2020instance,dong2020bi,chang2018rcaa} improve the feature discrimination with the guided query or surrounding persons.
\cite{dong2020bi} propose a bi-directional interaction network and utilize the cropped person patches as the guidance to reduce redundant context influences.
Considering the inconsistency of detection and Re-ID, \cite{chen2020norm} decompose the features of two subtasks in the polar coordinate system and propose a re-weighting method to focus on informative person areas.

In this work, we also adopt the end-to-end strategy to remain high efficiency. We use an external Re-ID model to provide strong guidance via multiple knowledge distillation and propose the spatial-invariant augmentation method to improve the robustness of Re-ID head 
against inaccurate detection results.

\myparagraph{Guidance from %
external networks.}
It is a common practice to improve the main network using the additional guidance from external networks~\cite{li2017mimicking,cheng2018revisiting,zhu2019deformable,hinton2015distilling,tung2019similarity,liu2019structured,yuan2020revisiting}.
Two recent works in person search~\cite{munjal2019knowledge,dong2020bi} are related to our approach. 
\cite{munjal2019knowledge} propose to distill the feature knowledge from a cumbersome model to a compact network on both detection and Re-ID subtasks, in which these two models solve the same person search task.
\cite{dong2020bi} %
use
the guidance from the cropped person patches to eliminate the context influence outside the bounding boxes.
Our proposed method aims to improve the discrimination of the Re-ID head by an 
independently trained Re-ID model.
The optimization is supervised from diverse knowledge distillation mechanisms.
Besides, our method makes the Re-ID head insensitive to inaccurate detection results via a spatial-invariant augmentation
scheme.

\section{Method}
In this section, we first conduct a comparative study to show that the Re-ID head is the main bottleneck in the end-to-end model (Section~\ref{sec:reid_importance}).
We then present the proposed person search framework,
as illustrated in Figure~\ref{fig:figure2}.
The overall structure can provide powerful guidances tailored for both input and output of the networks.
For output, we propose the diverse knowledge distillation scheme (Section~\ref{sec:external_guidance}) to extract multiple guidences accessed from the external Re-ID model, in which consisting of the probability-aware knowledge distillation (KD) and the relation-aware knowledge distillation.
As for input, a spatial-invariant augmentation method (Section~\ref{sec:spatial-invariant_augmentation}) is introduced to make the 
Re-ID head robust to the noisy detection results.

\subsection{Gap Between Two-Step and End-to-End Strategies}\label{sec:reid_importance}

Two first rows in Table~\ref{tab:Re-ID_importance} are the  results of end-to-end and two-step methods\footnote{For the two-step method, we train the detection and the Re-ID model individually. 
Bounding boxes are first produced by the detector and then passed through the Re-ID model to obtain the feature embeddings in the inference. For the end-to-end method, the joint model with the detection and Re-ID head is simultaneously trained and the feature embeddings are extracted via inputting the images to the joint model directly (details in Section 4.2).}.
It shows that the detection results in both settings are similar,
especially in the CUHK-SYSU dataset, \ie, 87.02\% \vs 87.23\% in the AP metric.
However, there is a large gap in their final matching results.
Specifically, the two-step strategy outperforms the end-to-end strategy by 5.5\%  and 9.9\% mAP on CUHK-SYSU and PRW datasets, respectively.
This motivates us to regard the Re-ID head as the main bottleneck of end-to-end person search models.
We propose our novel methods in the following sections to improve the capability of the Re-ID head and bridge the gap between the two kinds of strategies.

\begin{table}
\footnotesize
\caption{Component performance comparison between the end-to-end and two-step methods.
Model structures are described in Section~\ref{sec:ablation}.
}
\begin{center}
\setlength{\tabcolsep}{0.8mm}{
\begin{tabu} to 1.0\linewidth {l|X[c]|X[c]|X[c]|X[c]|X[c]|X[c]|X[c]|X[c]}
\hline
\multirow{3}{*}{Strategy}  & \multicolumn{4}{c|}{CUHK-SYSU} & \multicolumn{4}{c}{PRW} \\
\cline{2-9}
& \multicolumn{2}{c|}{Detection} & \multicolumn{2}{c|}{Re-ID} & \multicolumn{2}{c|}{Detection} & \multicolumn{2}{c}{Re-ID} \\
\cline{2-9}
 & AP & Recall & mAP & top-1 & AP & Recall & mAP & top-1 \\        
\hline
\hline
two-step & \multirow{1}{*}{87.02} & \multirow{1}{*}{89.69} & 91.97 & 93.14 & \multirow{1}{*}{92.39} & \multirow{1}{*}{94.08} & 51.85 & 85.37 \\
\hline
end-to-end & 87.23 & 90.02 & 86.47 & 87.55 & 90.85 & 92.93  & 41.91 & 82.11 \\
\hline
\OurExp & 87.87 & 90.62 & 93.09 & 94.24 & 90.71 & 92.71  & 50.51 & 87.07 \\
\hline
\end{tabu}}
\end{center}
\label{tab:Re-ID_importance}
\end{table}

\subsection{Diverse Knowledge Distillation}
\label{sec:external_guidance}

As indicated in Table~\ref{tab:Re-ID_importance}, the separate Re-ID model in the two-step strategy is able to accurately distinguish fine-grained pedestrians,
which inspires us to introduce an external well-trained Re-ID model (denoted as $\mathcal{M}$) to reinforce our Re-ID head (denoted as $\mathcal{H}$) in the end-to-end model.
The direct way is to apply knowledge distillation (KD) to transfer the informative knowledge from $\mathcal{M}$ to $\mathcal{H}$.
In addition to the traditional KD, \ie, probability-aware KD, we introduce a relation-aware KD scheme which consists of the pair-wise 
and the triplet-wise knowledge distillation.

\myparagraph{Probability-aware knowledge distillation.}
\label{sec:prob_guide}
Following the conventional KD~\cite{hinton2015distilling}, we employ a straightforward guidance to let head $\mathcal{H}$ mimic $\mathcal{M}$ by minimizing the Kullback-Leibler (KL) divergence between the predicted class probabilities of the two networks.
The output probability from $\mathcal{M}$ can be regarded as the soft target to serve as the informative guidance.
The loss function can be written as:
\begin{equation}
\centering
\begin{aligned}
{L}_{p} = \frac{1}{B}\sum_{i=1}^{B} \mathrm{KL}(\mathbf{p}_{i}^{\mathcal{H}}\parallel \mathbf{p}_{i}^{\mathcal{M}}),
\end{aligned}
\end{equation}    
where $\mathbf{p}_{i}^{\mathcal{H}}$ and $\mathbf{p}_{i}^{\mathcal{M}}$ denotes the probability output from 
$\mathcal{H}$ and 
$\mathcal{M}$ of the $i$-th sample in a mini-batch $B$. $\mathrm{KL}(\cdot)$ denotes KL divergence between two probabilities.

\begin{figure}[t!]
\centering
\includegraphics[trim =0mm 0mm 0mm 0mm, clip, width=0.9\linewidth]{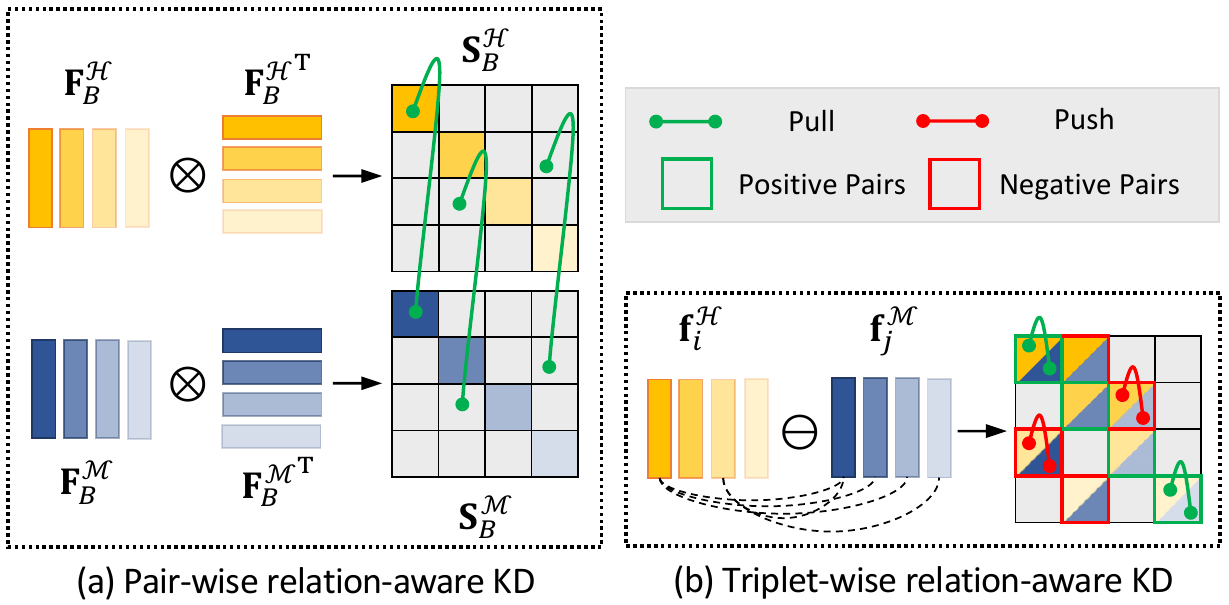}
\caption{Illustration of our proposed (a) pair-wise and (b) triplet-wise relation-aware KD.
$\mathcal{H}$ represents the Re-ID head in the end-to-end model, while $\mathcal{M}$ is the external Re-ID model.
Both two relation guidances push the Re-ID head to mimic the sample relationships of the Re-ID model.
}
\label{fig:figure3}
\end{figure}

\myparagraph{Relation-aware knowledge distillation.} 
Some works have already verified the success of discovering sample relationships in person Re-ID task~\cite{luo2019bag,he2020fastreid,hermans2017defense,zhang2020ordered,wang2020high,zhang2019self}.
It motivates us to transfer the informative relation knowledge among samples from $\mathcal{M}$ to $\mathcal{H}$. 
Inspired by~\cite{tung2019similarity,liu2019structured},
we present the \textit{pair-wise relation-aware knowledge distillation} to mimic the similarity distribution between sample pairs.
As shown in \xinyu{Figure~\ref{fig:figure3} (a)}, for a specific mini-batch, let $\mathbf{S}_{B}^{\mathcal{H}}$ and $\mathbf{S}_{B}^{\mathcal{M}}$ represent the similarity matrix among all sample pairs from the head $\mathcal{H}$ and model $\mathcal{M}$, respectively.
In detail, $\mathbf{S}_{B}^{\mathcal{H}}$ is produced by a row-wise L2 normalization on the outer product of the mini-batch embedding feature $\mathbf{F}_{B}^{\mathcal{H}}$:
\begin{equation}
\centering
\begin{aligned}
\mathbf{S}_{B}^{\mathcal{H}} \leftarrow \mathbf{F}_{B}^{\mathcal{H}} \otimes \mathbf{F}_{B}^{\mathcal{H}^{\mathrm{T}}}; \mathbf{S}_{B\left \{ i \right \}}^{\mathcal{H}} \leftarrow \mathbf{S}_{B\left \{ i \right \}}^{\mathcal{H}}/ || \mathbf{S}_{B\left \{ i \right \}}^{\mathcal{H}} ||_{2},
\end{aligned}
\end{equation}  
where $\left \{ i \right \}$ represents the $i$-th row in the similarity matrix.
The same process is applied on $\mathbf{S}_{B}^{\mathcal{M}}$ from the embedding feature $\mathbf{F}_{B}^{\mathcal{M}}$ in the model $\mathcal{M}$.
Then we apply the element-wise mean squared loss function on $\mathbf{S}_{B}^{\mathcal{H}}$ and $\mathbf{S}_{B}^{\mathcal{M}}$:
\begin{equation}
\centering
\begin{aligned}
{L}_{pr} = \frac{1}{B^2}||\mathbf{S}_{B}^{\mathcal{H}} - \mathbf{S}_{B}^{\mathcal{M}}||_{2}^{2},
\end{aligned}
\end{equation} 
In this way, the head $\mathcal{H}$ can capture the similar sample relationships as the model $\mathcal{M}$.

Motivated by recent contrastive learning algorithms~\cite{chen2020simple,he2020momentum,Tian2020Contrastive}, there exists a self supervision between differently generated views of the same data.
In our method, for a given data $x_i$, the feature embedding from $\mathcal{H}$ and $\mathcal{M}$, denoted as $\mathbf{f}_{i}^{\mathcal{H}}$ and $\mathbf{f}_{i}^{\mathcal{M}}$, can be viewed as a positive pair.
Instead, we treat any other samples within a mini-batch as negative examples.
The corresponding feature embedding $\mathbf{f}_{i}^{\mathcal{H}}$ and $\mathbf{f}_{j}^{\mathcal{M}}$ is thus a negative pair, where $i,j\in B$, $i\neq j$.
Following the above assumption, we propose the \textit{triplet-wise relation-aware knowledge distillation} to maximize the agreement of different feature representations of the same individual from $\mathcal{H}$ and $\mathcal{M}$, while minimizing that from different persons.
The loss function is defined as:
\begin{equation}
\centering
\begin{aligned}
{L}_{tr} = \frac{1}{B}\sum_{i=1}^{B}[m+||\mathbf{f}_{i}^{\mathcal{H}}-\mathbf{f}_{i}^{\mathcal{M}}||_2-\underset{\substack{i\neq j \\ i,j\in B}}{\mathrm{min}}||\mathbf{f}_{i}^{\mathcal{H}}-\mathbf{f}_{j}^{\mathcal{M}}||_2]_{+},
\end{aligned}
\label{eq:trg}
\end{equation} 
where $m$ is the margin between positive and negative pairs. Here, we mine the hardest negative sample $\mathbf{f}_{j}^{\mathcal{M}}$ within a mini-batch $B$ as in~\cite{hermans2017defense}.

We optimize the joint model with both of the probability-aware and the relation-aware KD to better capture sample probabilities and correlations from the external well-trained Re-ID model.
The whole objective function is presented as:
\begin{equation}
\centering
\begin{aligned}
L = L_{det} + \lambda (L_{reid} + \beta_{p}{L}_{p} + \beta_{pr}{L}_{pr} + \beta_{tr}{L}_{tr})
,
\end{aligned}
\label{eq:final_loss}
\end{equation}
where $L_{det}$ and $L_{reid}$ are the original detection and Re-ID loss functions. Here we use softmax cross-entropy loss
for $L_{reid}$. $\lambda$ acts as the weight to balance the detection and Re-ID losses. $\beta_{p}$, $\beta_{pr}$ and $\beta_{tr}$ are the loss weights for different types of guidances from $\mathcal{M}$, respectively.

\subsection{Spatial-Invariant Augmentation}\label{sec:spatial-invariant_augmentation}
In person Re-ID community, a major research line is to deal with the misalignment problem caused by occlusions, partial bodies or wrongly detected boxes~\cite{sun2019perceive,guo2019beyond,sun2018beyond,zhao2020not,miao2019pose}.
The goal of existing methods aims to learn a robust model to extract informative features from various misaligned persons.
In the person search task, the quality of Re-ID features is sensitive to the detection results to some extent.
To this end, we propose a spatial-invariant augmentation mechanism at both image level and feature level.

\myparagraph{Image-level spatial-invariant augmentation.} 
In \cite{chen2018person}, authors show 
that the expanded RoI works better for Re-ID model.
However, too much background will reduce the performance.
Instead of directly using the expanded images, 
we propose the image-level spatial-invariant augmentation (ISA) for model $\mathcal{M}$ to 
stably improve the capability of the external Re-ID guidance.
In detail, we expand the ground truth bounding boxes by a pixel length $\alpha$, resulting in loose boxes, in which they share the same centroids with the corresponding ground truths.
We then use the loose boxes with the random crop operation to train the Re-ID model, denoted as $\widetilde{\mathcal{M}}$.
Considering that there are multiple boxes with various scales, we need to impose an upper-bound restriction on the range of the random crop operation according to different box sizes.
In Re-ID, we usually add zero padding with $\Delta p$ pixels and then crop and resize the images to a fixed size $H\times W$.
Based on that, for a specific ground truth box with height $h$  and width $w$ , the expanded range of them can be adapted to $r_h=\mathrm{min}(2h\times \Delta p/{H},\alpha)$ and $r_w=\mathrm{min}(2w\times \Delta p/{W},\alpha)$ respectively.
$\mathrm{min}$ operation is to prevent the cropped boxes far from the ground truths.
In this way, the external model boosts the spatial-invariant ability by simulating inaccurate boxes.

\myparagraph{Feature-level spatial-invariant augmentation.}
In addition to ISA on the external $\widetilde{\mathcal{M}}$, we also improve the robustness of the Re-ID head $\mathcal{H}$ against inaccurate detection boxes.
Under the assumption that $\widetilde{\mathcal{M}}$ is good enough to recognize the identities even if the generated bounding boxes are not well aligned to ground truths,
we propose a feature-level spatial-invariant augmentation (FSA) to let $\mathcal{H}$ mimic the discriminative capability of $\widetilde{\mathcal{M}}$ on spatial variances.
Specifically, given a ground truth proposal, we perform the RoIAlign~\cite{he2017mask} in the range of the proportional pixels \wrt ~pixel range in ISA, \ie, $r_w / S$ and $r_h / S$.
Here, $S$ is the stride pixel of the current feature map \wrt the input image.
Consequently, there is an additional probability $\tilde{\mathbf{p}}_{i}^{\mathcal{H}}$ with the spatial variance for a given data $x_i$.

For better capturing the consistent information from $\widetilde{\mathcal{M}}$, we also extract the probability $\tilde{\mathbf{p}}_{i}^{\mathcal{M}}$ from $\widetilde{\mathcal{M}}$ by cropping the person patch
with the relative coordinate on the raw image \wrt the above spatial variance.
Then the objective function of the probability-aware KD in Section~\ref{sec:prob_guide} is re-defined as:
\begin{equation}
\centering
\begin{aligned}
{L}_{p} = \frac{1}{B}\sum_{i=1}^{B} [\mathrm{KL}(\mathbf{p}_{i}^{\mathcal{H}}\parallel \mathbf{p}_{i}^{\widetilde{\mathcal{M}}}) + \mathrm{KL}(\tilde{\mathbf{p}}_{i}^{\mathcal{H}}\parallel \tilde{\mathbf{p}}_{i}^{\widetilde{\mathcal{M}}}) ],
\end{aligned}
\label{eq:pg2}
\end{equation}    
where the original $\mathcal{M}$ is replaced by $\widetilde{\mathcal{M}}$ after ISA and 
${L}_{p}$ is updated from Eq.~\eqref{eq:final_loss} to Eq.~\eqref{eq:pg2}.
By this means, $\mathcal{H}$ is relatively insensitive to various detection results.

\section{Experiments}

\subsection{Evaluation Settings}
\myparagraph{Datasets.}
We carry out experiments on CUHK-SYSU and PRW datasets. 
CUHK-SYSU~\cite{xiao2017joint} is a large-scale dataset extracted from the street and movie snapshots. It consists of 18,184 images with 96,143 annotated bounding boxes with 8,432 labeled identities.
The training set contains 11,206 images with 5,532 identities, while the test set includes 6,978 gallery images and 2,900 query persons.
A set of protocols with gallery size ranging from 50 to 4000 is defined to evaluate the model scalability.
The gallery size is set to 100 by default without specificity.
PRW dataset~\cite{zheng2017person} is collected by six cameras at different locations on a university campus.
A total of 11,816 frames are annotated with 43,110 bounding boxes, among which 34,304 are assigned labels from 1 to 932, and the rest marked unknown identities.
The training set consists of 5,704 frames with 483 identities, and the test set contains 6,112 gallery images and 2,057 query images with 450 identities.

\myparagraph{Metrics.}
We use the standard evaluation metrics in person search~\cite{xiao2017joint}, \ie, cumulative matching characteristics (CMC) and mean average precision (mAP).
CMC is inherited from the person Re-ID problem showing the probabilities that at least one of the top-$K$ predicted bounding boxes overlaps with the ground truths with IoU greater or equal to 0.5.
mAP is inspired by object detection, where the average precision (AP) for each query is computed from the area under its precision-recall curve and then mAP is calculated by averaging APs across all queries.
For detection, we utilize Recall and Average Precision (AP) as the detection metrics following~\cite{chen2020norm}.

\subsection{Implementation Details}
\myparagraph{Model.}\label{sec:model_describe}
For the end-to-end model in our method, we adopt Faster R-CNN~\cite{ren2015faster} as our backbone network, in which ResNet-50~\cite{he2016deep} pretrained on ImageNet~\cite{deng2009imagenet} is used.
As in \cite{chen2020norm,dong2020bi}, the stem network has four residual blocks from conv1 to conv4.
A standard RPN is built on the stem network to generate candidate proposals, followed by the RoIAlign~\cite{he2017mask} operation to crop and reshape the proposals to $16\times 8$.
The detection head and the Re-ID head are two separate conv5 residual blocks without parameter sharing. 
The task-specific heads help alleviate the conflicts across tasks with a small overhead in computation.  
Following \cite{luo2019bag}, we adopt a batch-normalization layer before the last fully-connection layer.
The dimension of feature embeddings from the Re-ID head is set to 2048.
For the external Re-ID model, we employ a popular baseline~\cite{luo2019bag} based on the ResNet-50 backbone.

\begin{table}
\footnotesize
\caption{Efficacy of the proposed image-level augmentation (ISA) and the feature-level augmentation (FSA) on PRW dataset.
[X;Y] represents the cascaded operation of X and Y model.
Full notations are described in Section~\ref{sec:ablation}.
}
\begin{center}
\setlength{\tabcolsep}{0.9mm}{
\begin{tabu} to 0.95\linewidth {l|c|X[c]|X[c]|X[c]|X[c]}
\hline
\multirow{2}{*}{Strategy} & \multirow{2}{*}{Method} & \multicolumn{2}{c|}{Augmentation Type} & \multicolumn{2}{c}{PRW} \\
\cline{3-6}
 & & ISA & FSA & mAP & top-1 \\        
\hline
\hline
\multirow{4}{*}{two-step} & {[}D; R-1{]} & - & - & 51.85 & 85.37 \\
& {[}D; R-2{]} & \checkmark & - & 54.16  & 87.89 \\ 
\cline{2-6}
& {[}GT; R-1{]} & - & - & 54.46 & 86.63 \\
& {[}GT; R-2{]} & \checkmark & - & 56.78 & 89.30 \\
\hline
\hline
\multirow{4}{*}{end-to-end} & \OurExp-0 & - & - & 41.91 & 82.11 \\
 & \OurExp-1 & - & - & 49.44 & 85.85 \\
 & \OurExp-2 & \checkmark & - & 49.78 & 86.19 \\
 & \OurExp-3 & \checkmark & \checkmark & 50.15 & 86.53 \\
\hline
\end{tabu}}
\end{center}
\label{tab:SA}
\end{table}

\begin{table}
\footnotesize
\caption{Efficacy of different supervision guidance on PRW.}
\begin{center}
\setlength{\tabcolsep}{0.8mm}{
\begin{tabu} to 0.95\linewidth {l|c|X[c]|X[c]|X[c]|X[c]|X[c]}
\hline
\multirow{2}{*}{Strategy} & \multirow{2}{*}{Method} & \multicolumn{3}{c|}{Type} & \multicolumn{2}{c}{PRW} \\
\cline{3-7}
& & ${L}_{pr}$ & ${L}_{tr}$ & ${L}_{p}$ & mAP & top-1 \\        
\hline
\hline
two-step & {[}D; R-2{]} & - & - & - & 54.16  & 87.89 \\ 
\hline
\hline
\multirow{7}{*}{end-to-end} & \OurExp-0 & - & - & - & 41.91 & 82.11 \\
& \OurExp-4 & \checkmark & - & - & 46.00 & 84.59 \\
& \OurExp-5 & - & \checkmark & - & 45.54 & 83.37 \\
& \OurExp-6 & \checkmark & \checkmark & - & 47.18 & 84.83 \\
& \OurExp-7 & \checkmark & - & \checkmark & 50.18 & 86.78 \\
& \OurExp-8 & - & \checkmark & \checkmark & 50.46 & 86.58 \\
& \OurExp & \checkmark & \checkmark & \checkmark & \textbf{50.51} & \textbf{87.07} \\
\hline
\end{tabu}}
\end{center}
\label{tab:external_guidance}
\end{table}

\myparagraph{Training.}
We train all models using SGD optimizer with a momentum of 0.9 and a weight decay of $5\times 10^{-4}$.
Input images are resized to have at least 800 pixels on the short size and at most 1333 pixels on the long side.
We set the batch size to 4.
The initial learning rate is 0.001 and then multiplied by 0.1 after $3\times 10^4$ iterations.
The total iteration is $5\times 10^5$.
We set $\lambda=0.1$, $\beta_{p}=0.1\beta_{pr}=\beta_{tr}=1.0$ and $m=0.3$ in Eq.~\eqref{eq:trg} and $S$ to 16 in FSA.
For the Re-ID model, we use the softmax cross-entropy loss and the SGD optimizer.
We resize all the person patches to $H\times W=256\times 128$.
The batch size is set to 64.
The initial learning rate is $3.5\times 10^{-4}$ and the total training epoch is 240.
Besides, we set $\Delta p$ to 10 and $\alpha$ to 40 in the process of ISA.
In the inference stage, we only keep the joint model.

\begin{table}
\footnotesize
\caption{Person search results on CUHK-SYSU and PRW datasets. The gallery size of CUHK-SYSU is 100. $*$ denotes that QEEPS, QEEPS+ and APNet are from \cite{munjal2019query}, \cite{munjal2019knowledge} and \cite{zhong2020robust}.
}
\begin{center}
\setlength{\tabcolsep}{0.6mm}{
\begin{tabu} to 1.0\linewidth {c|l|c|c|c|c}
\hline
 & \multirow{2}{*}{Method} & \multicolumn{2}{c|}{CUHK-SYSU} & \multicolumn{2}{c}{PRW} \\
\cline{3-6}
 & & mAP & top-1 & mAP & top-1 \\        
\hline
\hline
\multirow{6}{*}{{\rotatebox{90}{Two-step}}} & MGTS~\cite{chen2018person} & 83.0 & 83.7 & 32.6 & 72.1 \\
 & CLSA~\cite{lan2018person} & 87.2 & 88.5 & 38.7 & 65.0 \\ 
 & RDLR~\cite{han2019re} & 93.0 & 94.2 & 42.9 & 70.2 \\ 
 & TCTS~\cite{wang2020tcts} & \textbf{93.9} & \textbf{95.1} & 46.8 & 87.5 \\ 
 & IGPN+PCB~\cite{dong2020instance} & 90.3 & 91.4 & 47.2 & 87.0 \\ 
 \cline{2-6} 
 & [D; R-2] (Ours two-step) & 93.60 & 94.72 & \textbf{54.16}  & \textbf{87.89} \\ 
\hline
\hline
\multirow{13}{*}{\rotatebox{90}{End-to-end}} & OIM~\cite{xiao2017joint} & 75.5 & 78.7 & 21.3 & 49.9 \\ 
 & IAN~\cite{xiao2019ian} & 76.3 & 80.1 & 23.0 & 53.1 \\ 
 & NPSM~\cite{liu2017neural} & 77.9 & 81.2 & 24.2 & 53.1 \\ 
 & RCAA~\cite{chang2018rcaa} & 79.3 & 81.3 & - & - \\ 
 & I-NeT~\cite{he2018end} & 79.5 & 81.5 & - & - \\ 
 & CTXGraph~\cite{yan2019learning} & 84.1 & 86.5 & 33.4 & 73.6 \\ 
& QEEPS$^{*}$ & 88.9 & 89.1 & 37.1 & 76.7 \\ 
 & {QEEPS+}$^{*}$ & 85.0 & 85.5 & 39.7 & 80.0 \\ 
 & APNet$^{*}$ & 88.9 & 89.3 & 41.9 & 81.4 \\ 
 & BINet~\cite{dong2020bi} & 91.5 & 92.4 & 45.3 & 81.7 \\ 
 & NAE+~\cite{chen2020norm} & 92.1 & 92.9 & 44.0 & 81.1 \\ 
 \cline{2-6} 
 & \OurExp-0 (Ours baseline) & 86.47 & 87.55 & 41.91 & 82.11 \\ 
 & \OurExp (Ours full) & \textbf{93.09} & \textbf{94.24} & \textbf{50.51} & \textbf{87.07} \\ 
 \hline
\end{tabu}}
\end{center}
\label{tab:sota}
\vspace{-3mm}
\end{table}

\subsection{Ablation Study}\label{sec:ablation}
For simplicity, we first introduce the following variants:
\begin{itemize}
\itemsep -.051cm
\item D: the detection model only with $L_{det}$ in Eq.~\eqref{eq:final_loss}.
\item R-1: the Re-ID model $\mathcal{M}$ as described in Section~\ref{sec:model_describe}. The model is trained with person patches with labeled ground truth bounding boxes.
\item R-2: adding the proposed ISA on person patches compared with R-1. 
\item \OurExp-0: the plain model trained in the end-to-end strategy. Details are described in Section~\ref{sec:model_describe}. $L_{det}$ and $L_{reid}$ are used for optimization in Eq.~\eqref{eq:final_loss}.
\item \OurExp-1: adding the probability-aware knowledge distillation (KD) from R-1 on \OurExp-0. $L_{det}$, $L_{reid}$ and ${L}_{p}$ are loss functions in Eq.~\eqref{eq:final_loss}.
\item \OurExp-2: adding the probability-aware KD from R-2 on \OurExp-0.
\item \OurExp-3: adding the proposed FSA on \OurExp-2, \ie, replacing ${L}_{p}$ with Eq.~\eqref{eq:pg2} in Eq.~\eqref{eq:final_loss}. %
\item \OurExp-4 / -5: adding the proposed pair-wise / triplet-wise relation-aware KD from R-2 on \OurExp-0. $L_{det}$, $L_{reid}$ and ${L}_{pr}$ / ${L}_{tr}$ are loss functions in Eq.~\eqref{eq:final_loss}.
\item \OurExp-6: adding both pair-wise and triplet-wise relation-aware KD from R-2 on \OurExp-0 simultaneously. $L_{det}$, $L_{reid}$, ${L}_{pr}$ and ${L}_{tr}$ are loss functions in Eq.~\eqref{eq:final_loss}.
\item \OurExp-7 / -8: adding probability-aware KD in Eq.~\eqref{eq:pg2} on \OurExp-4 / -5. ${L}_{p}$ is added as the loss function in Eq.~\eqref{eq:final_loss}.
\item \OurExp: adding probability-aware KD in Eq.~\eqref{eq:pg2} on \OurExp-6, which is the whole setting of our method.
\end{itemize}

\myparagraph{Effectiveness of ISA and FSA.}
Table~\ref{tab:SA} shows the ablation study results on PRW dataset.
The results ([D; R-1] \vs [D; R-2] and \OurExp-1 \vs \OurExp-2) demonstrate the efficacy of our image-level spatial-invariant augmentation (ISA).
It shows that ISA leads to 2.31\% mAP improvement in the two-step setting.
When applying the Re-ID model with ISA as the external guidance in our method, there are 0.34\% mAP and 0.34\% top-1 improvements.
It is because the Re-ID model with ISA can provide a robust external guidance when detection results are inaccurate. 
Moreover, in \OurExp-3, we apply the feature-level spatial-invariant augmentation (FSA) on \OurExp-2, resulting in further improvement of the performance to 50.15\% mAP and 86.53\% top-1.
It is because that the Re-ID head can effectively learn the spatial invariance with the help of FSA,
so that the influences from the detection head can be suppressed to some extent.

\myparagraph{Effectiveness of different knowledge distillation schemes.}
We evaluate the impact of different KD schemes as described in Section~\ref{sec:external_guidance}.
Results are performed on PRW dataset and shown in Table~\ref{tab:external_guidance}, from which we can observe that:
i) Our Re-ID head successfully learns the discriminative ability under the guidance from the external Re-ID model. 
With all three types of the supervision guidances, our method improves 8.6\% mAP and 4.96\% top-1, which is close to the results of two-step strategy.
ii) Either pair-wise or triplet-wise relation-aware KD can improve the performance and the result is further improved with both of them.
We believe that it is because our Re-ID head well captures reliable sample relationships provided by the Re-ID model which improves the feature discrimination.

\begin{figure}[t!]
\centering
\includegraphics[trim =0.1mm 0mm 0mm 0mm, clip, width=1.01\linewidth]{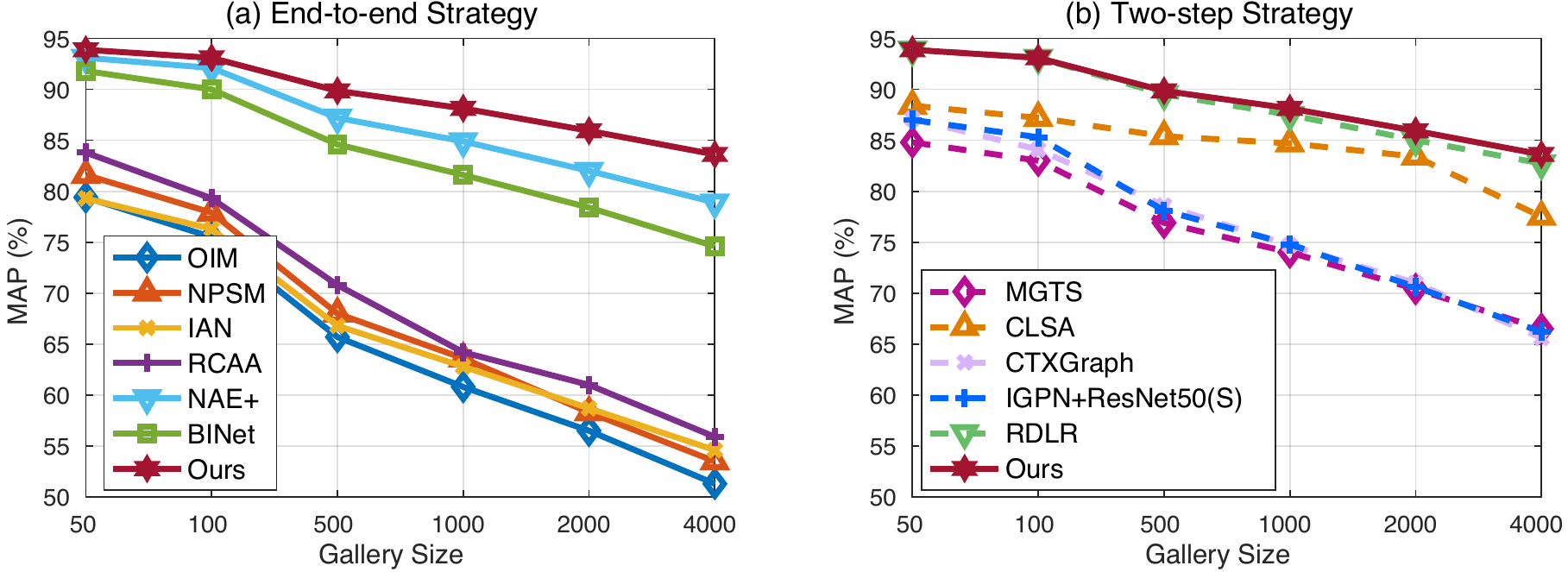}
\caption{Results on CUHK-SYSU with varying gallery sizes. 
}
\label{fig:gallery_size}
\end{figure}

\begin{figure*}[t!]
\centering
\includegraphics[trim =0mm 0mm 0mm 0mm, clip, width=0.95\linewidth]{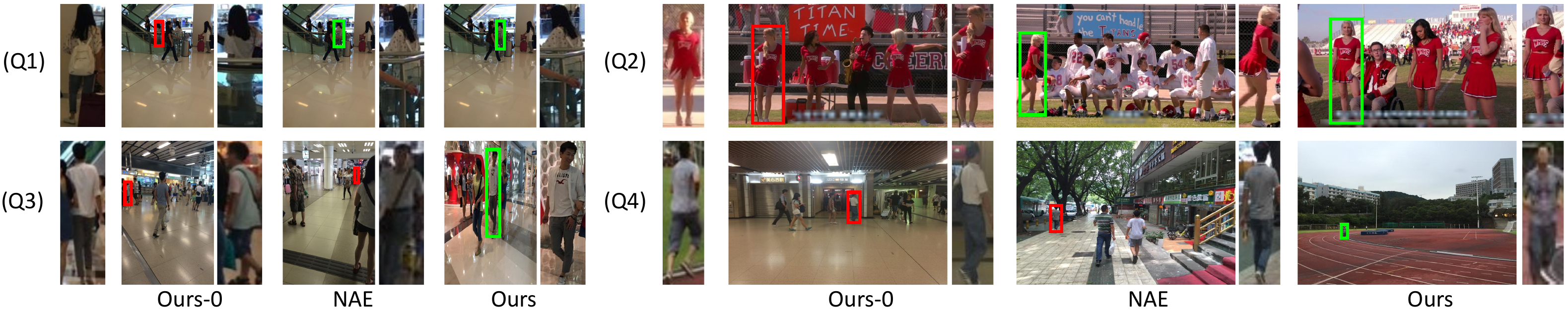}
\caption{Sample visual results.
For each query 'Q', we show the top-1 search result by our baseline (\OurExp-0), NAE~\cite{chen2020norm} and the our whole method (\OurExp). 
\textcolor{green}{Green} and \textcolor{red}{red} denote the correct and wrong results respectively. 
}
\label{fig:visual}
\end{figure*}

\subsection{Comparison with State-of-the-art Methods}
Following the evaluation setting in~\cite{xiao2017joint,zheng2017person,chen2020norm}, we compare our proposed method with state-of-the-art methods on person search.
As shown in Table~\ref{tab:sota},
all methods are categorized into two parts, \ie, two-step strategies in the upper block and the end-to-end methods in the lower block.

\myparagraph{Results on CUHK-SYSU.}
As shown in Table~\ref{tab:sota}, our method achieves the best results in the end-to-end category.
Compared with NAE+~\cite{chen2020norm} that decomposes the detection and Re-ID subtasks via a norm-aware embedding, our method focuses on the improvement of the Re-ID head guided by the external Re-ID model and results in the superior performance (93.09\% \vs 92.1\% mAP).
Our method also outperforms another prior best method BINet~\cite{dong2020bi} by 1.59\% mAP.
Instead of alleviating the negative context information as in~\cite{dong2020bi},
our method aims to learn spatial-invariant features that are insensitive to inaccurate detection results.
Moreover, our proposed relation-aware KD explores the sample relationships, which is important to the Re-ID task.
From Figure~\ref{fig:visual}, top-1 search results from the baseline, NAE~\cite{chen2020norm} and our method are visualized.

We also evaluate the scalability of different methods with various gallery sizes,
\ie, \{50, 100, 500, 1000, 2000, 4000\}.
As shown in Figure~\ref{fig:gallery_size}, all methods degrade the performance monotonically as the gallery size increases.
It shows that person search in large search scopes remains a challenging task as it involves more distractors.  
Nevertheless, our method outperforms all approaches, especially under large gallery sizes.
There is only a relatively small drop in mAP when the gallery size changes from 50 to 4000, which demonstrates the scalability of our approach.
Interestingly, our method achieves comparable results and even surpasses all two-step methods with the gallery size increasing, which shows that
our method can improve the model discrimination and obtain more spatial-invariant features.

\begin{table}
\footnotesize
\caption{Running time (ms) comparison.
[D; R-2] stands for the two-step model cascaded by detection and Re-ID.
}
\begin{center}
\setlength{\tabcolsep}{1.0mm}{
\begin{tabu} to 0.9\linewidth {l|X[c]|X[c]|X[c]|X[c]}
\hline
\multirow{2}{*}{Image Size} & \multicolumn{3}{c|}{End-to-end} & \multicolumn{1}{c}{Two-step} \\ 
\cline{2-5}
 & \OurExp & NAE & BINet & {[}D; R-2{]} \\ 
 \hline
900x1500 & 124  & 141 & 80 & 175 \\ 
\hline
\end{tabu}}
\end{center}
\label{tab:time}
\end{table}

\begin{table}
\footnotesize
\caption{Effect of different external Re-ID models on PRW.
}
\begin{center}
\setlength{\tabcolsep}{1.0mm}{
\begin{tabu} to 0.9\linewidth {l|X[c]|X[c]}
\hline
Re-ID model & mAP & top-1 \\ 
 \hline
\OurExp & 50.51  & 87.07 \\ 
 \hline
w/o random erasing~\cite{zhong2020random} & 50.19  & 86.58 \\
w/o BNNeck~\cite{luo2019bag} & 47.83  & 85.03 \\
\hline
\end{tabu}}
\end{center}
\label{tab:differnt_teacher}
\end{table}

\myparagraph{Results on PRW.}
As shown in Table~\ref{tab:sota}, our baseline model is already competitive with other end-to-end methods.
This indicates that it would be better to separate the detection and Re-ID heads in the end-to-end model without parameter sharing.
Overall, our method surpasses all the methods both in the end-to-end and two-step strategy except TCTS~\cite{wang2020tcts} at top-1.
Notably, it outperforms the best end-to-end method BINet~\cite{dong2020bi} by 5.21\% mAP, showing the superiority of our method.

\myparagraph{Running time.}
We report the speed of different approaches in Table~\ref{tab:time}.
All experiments are conducted on a GeForce GTX 1080 Ti machine.
For fairness, the size of input images is set to $900\times 1500$.
We reproduce the result of NAE~\cite{chen2020norm} on the same machine, while reporting the original result of BINet~\cite{dong2020bi} since its code is not available.
Table~\ref{tab:time} shows that our method is faster than NAE.
Note that our method can achieve the same efficiency when applying the same backbone as BINet, as no additional parameter is introduced.
Moreover, our method is more efficient than the two-step strategy (\ie, {[}D; R-2{]}).

\myparagraph{Discussion.} In order to show the effectiveness of our method clearly, we evaluate the effect of different external Re-ID models. As shown in Table~\ref{tab:differnt_teacher}, when removing the important schemes in the Re-ID model, \ie, random erasing~\cite{zhong2020random} and BNNeck~\cite{luo2019bag},
the performance is slightly lower than the full model yet outperforms all the 
recent 
methods.
We also explore the detection results influenced by our method in Table~\ref{tab:Re-ID_importance}. Compared with the baseline, our method has limited influence on the detection part.

\section{Conclusion}
In this paper, we identify that the Re-ID head is the main bottleneck in the end-to-end person search model.
To this end, we propose a strong person search network to improve the Re-ID head with an external Re-ID model.
We provide guidance on both the output and input of the model.
For the model output, we design a diverse knowledge distillation, consisting of probability-aware and relation-aware KD, to let the Re-ID head mimic the well-trained Re-ID model.
For the model input, we develop an image-level and feature-level spatial-invariant augmentations to make the Re-ID head insensitive to inaccurate detection results.
Only the end-to-end model is needed during inference so that there is no additional computation.
Extensive experiments show the effectiveness of our methods.

\clearpage

\section*{Acknowledgments}
MY is the corresponding author. 
XZ and MY were in part
supported by the Shanghai Natural Science Foundation under grant no.\  18ZR1442600, the National Natural Science Foundation of China under grant no.\  62073244 and the Shanghai Innovation Action Plan under grant no.\  20511100500.
XZ's contribution was made when visiting The University of Adelaide.
CS and his employer received no financial support for the research, authorship, and/or publication of this article.

\small
\bibliography{references}

\end{document}